\documentclass[11pt]{article}
\usepackage{acl2015}
\usepackage{times}
\usepackage{url}
\usepackage{latexsym}

\usepackage{qtree}
\usepackage{tree-dvips}
\usepackage{tikz}
\usepackage{tikz-qtree}

\usepackage{subfigure}
\usepackage{ccg}
\usepackage{amsmath}

\usepackage{multirow}
\usepackage{latexsym}
\usepackage{url}
\usepackage{epsfig}
\usepackage{subfigure}
\usepackage{array}
\usepackage{hhline}
\usepackage{pbox}

\newenvironment{smalign}{\par\nobreak\small\noindent\align}{\endalign}

\title{Learning the Semantics of Manipulation Action}

\author{Yezhou Yang\textsuperscript{$\dagger$} \and Yiannis Aloimonos\textsuperscript{$\dagger$} \and Cornelia Ferm\"{u}ller\textsuperscript{$\dagger$} \and Eren Erdal Aksoy\textsuperscript{$\ddagger$} \\
  \textsuperscript{$\dagger$} UMIACS, University of Maryland, College Park, MD, USA \\
  {\tt \{yzyang, yiannis, fer\}@umiacs.umd.edu} \\
  \textsuperscript{$\ddagger$} Karlsruhe Institute of Technology, Karlsruhe, Germany \\
  {\tt eren.aksoy@kit.edu } \\}

\date{\today}

\begin{document}
\maketitle
\begin{abstract}

In this paper we present a formal computational framework for modeling manipulation actions. The introduced formalism leads to semantics of manipulation action and has applications to both  observing and understanding human manipulation actions as well as executing them with a robotic mechanism (e.g. a humanoid robot). It is based on a Combinatory Categorial Grammar. The goal of the introduced framework is to:
(1) represent manipulation actions with both syntax and semantic parts, where the semantic part employs  $\lambda$-calculus;
(2) enable a probabilistic semantic parsing schema to learn the  $\lambda$-calculus representation of manipulation action from an annotated action corpus of videos;
(3) use (1) and (2) to develop a system that visually observes manipulation actions and understands their meaning while it can reason beyond observations using propositional logic and axiom schemata.
The experiments conducted on a public available large manipulation action dataset validate the theoretical framework and our implementation.
\end{abstract}

\section{Introduction}

%Problem statement (what do you want to solve) 
%goal of the system (final output) 
Autonomous robots will need to learn the actions that humans perform. They will need to recognize these actions when they see them and they will need to perform these actions themselves. This requires a formal system to represent the action semantics. This representation needs to store the semantic information about the actions, be encoded in a machine readable language, and inherently be in a programmable fashion in order to enable reasoning beyond observation. A  formal representation of this kind  has a variety of other applications such as intelligent manufacturing, human robot collaboration, action planning and  policy design, etc.

In this paper,  we are concerned with manipulation actions, that is actions performed by agents (humans or robots) on objects,  resulting in some physical change of the object. However most of the current AI systems require manually defined semantic rules. In this work, we propose a computational linguistics framework, which is based on probabilistic semantic parsing with Combinatory Categorial Grammar (CCG), to learn manipulation action semantics (lexicon entries) from annotations. We later show that this learned lexicon is able to make our system reason about manipulation action goals beyond just observation. Thus the intelligent system can not only imitate human movements, but also imitate action goals.

%current situation (difficulty) 
Understanding actions by observation and executing them are generally considered as dual problems for intelligent agents. The sensori-motor bridge connecting the two tasks is essential, and a great amount of attention in AI, Robotics as well as Neurophysiology has been devoted to investigating it. Experiments conducted on primates have discovered that certain neurons, the so-called mirror neurons, fire during both observation and execution of identical manipulation tasks \cite{rizzolatti2001neurophysiological,gazzola2007anthropomorphic}. This suggests that  the same process is involved in both the observation and execution of actions. From a functionalist point of view, such a process should be able to first build up a semantic structure from observations, and then the decomposition of that same structure should occur when the intelligent agent executes commands.

Additionally, studies in linguistics \cite{steedman2002plans} suggest that the language faculty develops in humans as a direct adaptation of a more primitive apparatus for planning goal-directed action in the world by composing affordances of tools and consequences of actions. It is this more primitive apparatus that is our major interest in this paper. Such an apparatus is composed of a ``syntax part'' and a ``semantic part''. In the syntax part, every linguistic element is categorized as either a function or a basic type, and is associated with a syntactic category which either identifies it as a function or a basic type. In the semantic part, a semantic translation is attached following the syntactic category explicitly.

Combinatory Categorial Grammar (CCG) introduced by \cite{steedman2000syntactic} is a theory that can be used to represent such structures with a small set of combinators such as functional application and type-raising. What do we gain though from such a formal description of action? This is similar to asking what one gains from a formal description of language as a generative system. Chomsky’s contribution to language research was exactly this: the formal description of language through the formulation of the Generative and Transformational Grammar \cite{Chomsky1957}. It revolutionized language research opening up new roads for the computational analysis of language, providing researchers with common, generative language structures and syntactic operations, on which language analysis tools were built. A grammar for action would contribute to providing a common framework of the syntax and semantics of action, so that basic tools for action understanding can be built, tools that researchers can use when developing action interpretation systems, without having to start development from scratch. The same tools can be used by robots to execute actions.

In this paper, we propose an approach for learning the semantic meaning of manipulation action through a probabilistic semantic parsing framework based on CCG theory. For example, we want to learn from an annotated training action corpus that the action ``Cut'' is a function which has two arguments: a subject and a patient. Also, the action consequence of ``Cut'' is a separation of the patient. Using formal logic representation, our system  will learn the semantic representations of ``Cut'': 
\begin{smalign}
Cut := & (AP\backslash NP)/NP : \lambda x. \lambda y. cut(x,y)  \rightarrow divided(y) \nonumber 
\end{smalign}
Here $cut(x,y)$ is a primitive function. We will further introduce the representation in Sec.~\ref{sec:mas}. Since our action representation is in a common calculus form, it enables naturally further logical reasoning beyond visual observation.

\begin{figure}[!ht]
\centering
\includegraphics[width=\columnwidth]{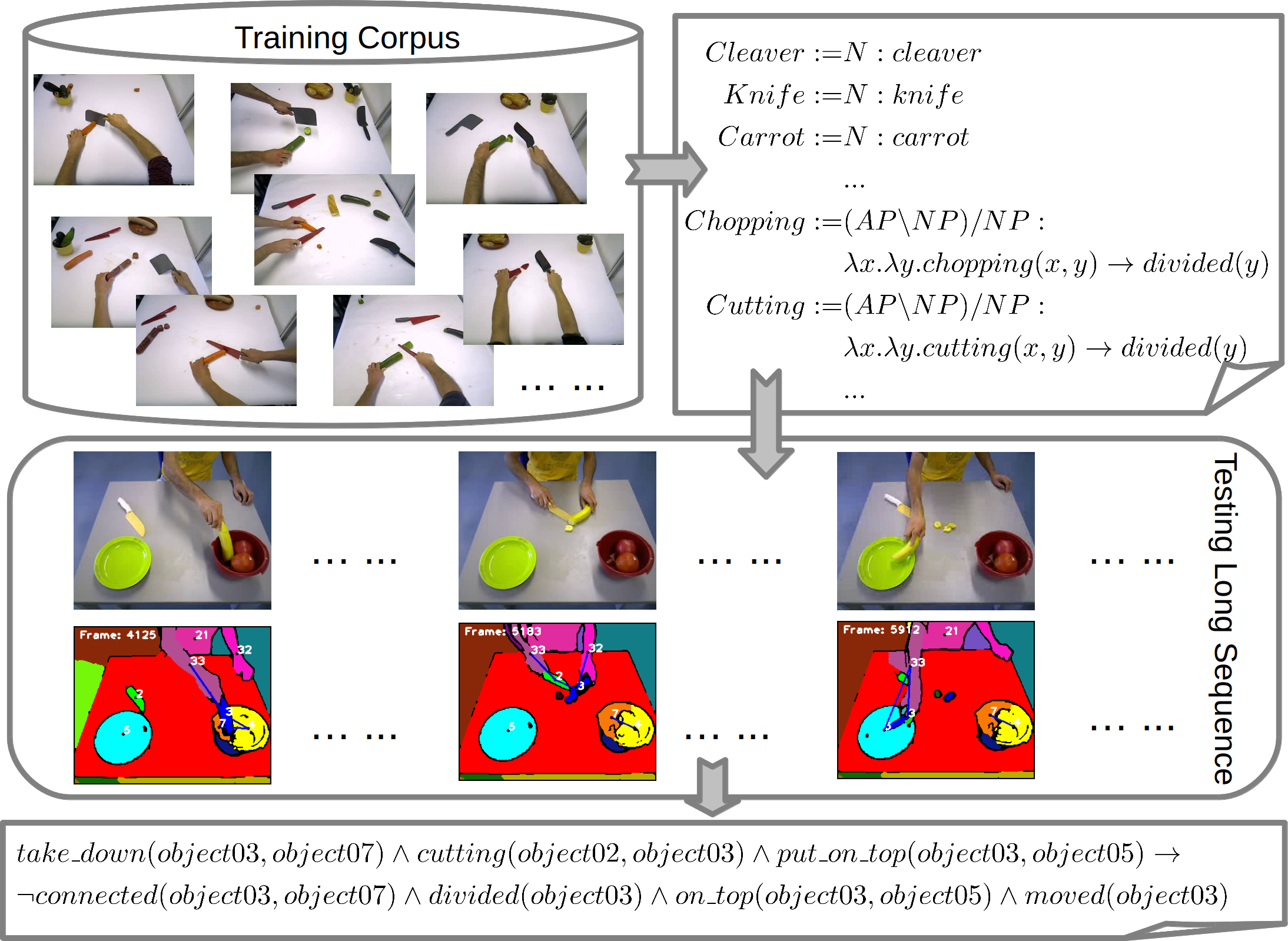}
\caption{A CCG based semantic parsing framework for manipulation actions.}
\label{fig:frame}
\end{figure}

The advantage of our approach is twofold: 1) Learning semantic representations from annotations helps an intelligent agent to enrich automatically its own knowledge about  actions; 2) The formal logic representation of the action could be used to infer the object-wise consequence after a certain manipulation, and can also be used to plan a set of actions to reach a certain action goal.
We further validate our approach on a large publicly available manipulation action dataset (MANIAC) from \cite{aksoytamosiunaitewoergoetter2014}, achieving promising experimental results.
Moreover, we believe that our work, even though it only considers the domain of manipulation actions, is also a promising example of a more closely intertwined computer vision and computational linguistics system. The diagram in Fig.\ref{fig:frame} depicts the framework of the  system.

\section{Related Works}

{\bf Reasoning beyond appearance: }The very small number of works in computer vision, which aim to reason beyond appearance models, are also related to this paper. \cite{xie2013inferring} proposed that beyond state-of-the-art computer vision techniques, we could possibly infer implicit information (such as functional objects) from video, and they call them ``Dark Matter'' and ``Dark Energy''. \cite{yezhou2013cvpr} used stochastic tracking and graph-cut based segmentation to infer manipulation consequences beyond appearance. \cite{joo2014visual} used a ranking SVM to predict the persuasive motivation (or the intention) of the photographer who captured an image. More recently, \cite{pirsiavash2014inferring} seeks to infer the motivation of the person in the image by mining knowledge stored in a large corpus using natural language processing techniques. Different from these fairly general investigations about reasoning beyond appearance, our paper seeks to learn manipulation actions semantics in logic forms through CCG, and further infer hidden action consequences beyond appearance through reasoning.

{\bf Action Recognition and Understanding:} Human activity recognition and understanding has  been studied heavily  in Computer Vision recently, and there is a large range of applications for this work in areas like human-computer interactions, biometrics, and video surveillance. Both  visual recognition methods, and the non-visual description methods using motion capture systems have been used. A few good surveys of the former can be found in \cite{moeslund2006survey} and \cite{turaga2008machine}. %and \cite{gavrila1999visual}. 
%The approach of  most  visual recognition methods is to learn from a large amount of spatio-temporal points what an action looks like (\cite{laptev2005space}, \cite{dollar2005behavior}, \cite{wang2007learning}, \cite{willems2008efficient}). 
Most of the focus has been  on recognizing single human actions like walking, jumping, or running etc. \cite{ben2002human,yilmaz2005actions}. 
Approaches to more  complex actions have employed parametric approaches, such as HMMs \cite{kale2004identification} to learn the transition between feature representations in individual frames e.g. \cite{saisan2001dynamic,chaudhry2009histograms}. %hu2000extraction,
More recently,  \cite{aksoy2011learning,aksoytamosiunaitewoergoetter2014} proposed a semantic event chain (SEC) representation to model and learn the semantic segment-wise relationship transition from spatial-temporal video segmentation. %This is the primary thrust of research in computer vision for such recognition techniques.

There also have been many syntactic approaches to human activity recognition which used the concept of context-free grammars, because such grammars provide a sound theoretical basis for modeling structured processes. Tracing back to the middle 90's, \cite{brand1996understanding} used a grammar to recognize disassembly tasks that contain hand manipulations. \cite{ryoo2006recognition} used the context-free grammar formalism to recognize composite human activities and multi-person interactions. It is a two level hierarchical approach where the lower-levels are composed of HMMs and Bayesian Networks while the higher-level interactions are modeled by CFGs. To deal with errors from low-level processes such as tracking, stochastic grammars such as stochastic CFGs were also used \cite{ivanov2000recognition,moore2002recognizing}. More recently, \cite{kuehne2014language} proposed to model goal-directed human activities using Hidden Markov Models and treat sub-actions just like words in speech. These works proved that grammar based approaches are practical in activity recognition systems, and shed insight onto  human manipulation action understanding. However, as %manipulation actions constitute a dual problem (interpretation and generation)
mentioned, thinking about manipulation actions  solely from the viewpoint of recognition has obvious limitations. In this work we adopt principles from CFG based activity recognition systems, with extensions to a CCG grammar that accommodates not only the hierarchical structure of human activity but also action semantics representations. It enables the system to serve as the core parsing engine for both manipulation action recognition and execution.

{\bf Manipulation Action Grammar: }As mentioned before, \cite{chomsky1993lectures} suggested that a minimalist generative grammar, similar to  the one of human language, also exists for action understanding and execution. The works closest  related to this paper are \cite{pastra2012minimalist,summersusing,guha2013plan}. \cite{pastra2012minimalist} first discussed  a Chomskyan grammar for understanding complex actions as a theoretical concept,  and \cite{summersusing} provided an implementation  of such a grammar using as perceptual input only objects. More recently, \cite{YezhouACS} proposed a set of context-free grammar rules for manipulation action understanding, and \cite{yang2015robot} applied it on unconstrained instructional videos. However, these approaches only consider the syntactic structure of manipulation actions without coupling semantic rules using $\lambda$ expressions, which limits the capability of doing reasoning and prediction.

{\bf Combinatory Categorial Grammar and Semantic Parsing:} CCG based semantic parsing originally  was used mainly to  translate natural language sentences to their desired semantic representations as  $\lambda$-calculus formulas \cite{zettlemoyerlearning,zettlemoyer2007online}. \cite{mooney2008learning} presented a framework of grounded language acquisition: the interpretation of language entities into semantically informed structures in the context of perception and actuation. The concept has been applied successfully in tasks  such as robot navigation \cite{matuszek2012joint}, forklift operation \cite{tellex2014learning} and of  human-robot interaction \cite{matuszek2014learning}. In this work, instead of grounding natural language sentences directly, we ground information obtained from visual  perception  into semantically informed structures, specifically in the domain of manipulation actions.

\section{A CCG Framework for Manipulation Actions}
\label{sec:mas}
Before we dive into the semantic parsing of  manipulation actions, a brief introduction to the Combinatory Categorial Grammar framework in Linguistics is necessary. We will only introduce related concepts and formalisms. For a complete background reading, we would like to refer readers to \cite{steedman2000syntactic}. We will first give a brief introduction to CCG and then introduce a fundamental combinator, i.e., functional application. The introduction is followed by examples to show how the combinator is  applied to parse actions.

\subsection{Manipulation Action Semantics}

The semantic expression in our representation of manipulation actions uses a typed $\lambda$-calculus language. The formal system has two basic types: entities and functions. Entities in manipulation actions are Objects or Hands, and functions are the Actions. Our $lambda$-calculus expressions  are formed from the following items:

\textbf{Constants}: Constants can be either entities or functions. For example, \textit{Knife} is an entity (i.e., it is of type N) and \textit{Cucumber} is an entity too (i.e., it is of type N). \textit{Cut} is an action function that maps entities to entities. When the event \textit{Knife Cut Cucumber} happened, the expression \textit{cut(Knife, Cucumber)} returns an entity of type AP, aka. Action Phrase. Constants like \textit{divided} are status functions that map entities to truth values. The expression $divided(cucumber)$ returns a true value after the event (\textit{Knife Cut Cucumber}) happened.  

\textbf{Logical connectors}: The $\lambda$-calculus expression has logical connectors like conjunction ($\wedge$), disjunction ($\vee$), negation($\neg$) and implication($\rightarrow$).

For example, the expression 
\begin{smalign}
connected(tomato, cucumber) \wedge \nonumber \\
divided(tomato) \wedge divided(cucumber) \nonumber
\end{smalign}
represents the joint status that the sliced \textit{tomato} merged with the sliced \textit{cucumber}. It can be regarded as a simplified goal status for ``making a cucumber tomato salad''. The expression $\neg connected(spoon, bowl)$ represents the status after the \textit{spoon} finished stirring the \textit{bowl}. 
\begin{smalign}
\lambda x.  cut(x,cucumber)  \rightarrow divided(cucumber) \nonumber
\end{smalign} 
represents that if the \textit{cucumber} is cut by $x$, then the status of the \textit{cucumber} is divided.

\textbf{$\lambda$ expressions}: $lambda$ expressions represent functions with unknown arguments. For example, $\lambda x. cut(knife, x)$ is a function from entities to entities, which is of type NP after any entities of type N that is cut by \textit{knife}. 

\subsection{Combinatory Categorial Grammar}
\label{sec:ccg}

The semantic parsing formalism underlying our framework for manipulation actions is that of combinatory categorial grammar (CCG) \cite{steedman2000syntactic}. A CCG specifies one or more logical forms for each element or combination of elements for manipulation actions. In our formalism, an element of Action is associated with a syntactic ``category'' which identifies it as functions, and specifies the type and directionality of their arguments and the type of their result. For example, action ``Cut'' is a function from patient object phrase (NP) on the right into predicates, and into functions from subject object phrase (NP) on the left into a sub action phrase (AP):
\begin{smalign}
Cut := (AP\backslash NP)/NP .
\label{eq:grasp}
\end{smalign}

\begin{figure}[!ht]
    \centering
      \begin{tikzpicture}[scale=0.8]
    \Tree  [.AP [.NP [.N Knife ] ] [.AP [.A [.Cut ] ] [.NP [.N Cucumber ] ] ] ]
    \end{tikzpicture}
    \label{fig:ex1}
    \caption{Example of conventional tree structure.}
\end{figure}
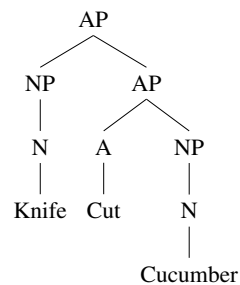

As a matter of fact, the pure categorial grammar is a conext-free grammar presented in the accepting, rather than the producing direction. The expression (\ref{eq:grasp}) is just an accepting form for Action ``Cut'' following the context-free grammar. While it is now convenient to write derivations as follows, they are equivalent to conventional tree structure derivations in Figure.~\ref{fig:ex1}.
\begin{center}
\small
\deriv{3}{
{\rm Knife}&{\rm Cut}& {\rm Cucumber}\\
\uline{1}& \uline{1}    & \uline{1}\\
\it N   &  &\it  N\\
\uline{1}& \uline{1}    & \uline{1}\\
\it NP   &\it (AP\bs NP)/NP &\it  NP\\
         &     \fapply{2}\\
         & \mc{2}{\it AP\bs NP}\\
\bapply{2}&\\
\mc{2}{\it AP}&
}
\end{center} 

The semantic type is encoded in these categories, and their translation can be made explicit in an expanded notation. Basically a $\lambda$-calculus expression is attached with the syntactic category. A colon operator is used to separate syntactical and semantic expressions, and the right side of the colon is assumed to have lower precedence than the left side of the colon. Which is intuitive as any explanation of manipulation actions should first obey syntactical rules, then semantic rules. Now the basic element, Action ``Cut'', can be further represented by:
\begin{smalign}
Cut := & (AP\backslash NP)/NP : \lambda x. \lambda y. cut(x,y)  \rightarrow divided(y) .\nonumber 
\end{smalign} 
$(AP\backslash NP)/NP$ denotes  a phrase of type $AP$, which requires an element of type $NP$ to specify what object was cut, and requires another element of type $NP$ to further complement what effector initiates the cut action. $\lambda x. \lambda y. cut(x,y)$ is the $\lambda$-calculus representation for this function. Since the functions are closely related to the state update, $\rightarrow divided(y)$ further points out the status expression after the action was performed.

A CCG system has a set of combinatory rules which describe how adjacent syntatic categories in a string can be recursively combined. In the setting of manipulation actions, we want to point out that similar combinatory rules are also applicable. Especially the functional application rules are essential in our system.

\subsection{Functional application}

The functional application rules with semantics can be expressed in the following form:
\begin{eqnarray}
A/B:f ~~~ B:g => A:f(g)  \label{eq:app1}\\
B:g ~~~ A\bs B:f => A:f(g)  \label{eq:app2}
\end{eqnarray}
Rule.~(\ref{eq:app1}) says that a string with type $A/B$ can be combined with a right-adjacent string of type $B$ to form a new string of type $A$. At the same time, it also specifies how the semantics of the category $A$ can be compositionally built out of the semantics for $A/B$ and $B$. Rule.~(\ref{eq:app2}) is a symmetric form of Rule.~(\ref{eq:app1}).

In the domain of manipulation actions, following derivation is an example CCG parse. This parse shows how the system can parse an observation (``Knife Cut Cucumber'') into a semantic representation ($cut(knife,cucumber) \rightarrow divided(cucumber)$) using the functional application rules. 
\begin{center}
\small
\deriv{3}{
{\rm Knife}&{\rm Cut}& {\rm Cucumber}\\
\uline{1}& \uline{1}    & \uline{1}\\
\it N   & &\it  N\\
\uline{1}& \uline{1}    & \uline{1}\\
\it NP   &\it (AP\bs NP)/NP &\it  NP\\
\it knife   &\it \lambda x. \lambda y. cut(x,y) &\it  cucumber\\
\it knife   &\it \rightarrow divided(y) &\it  cucumber\\
         &     \fapply{2}\\
         & \mc{2}{\it AP\bs NP}\\
         & \mc{2}{\it \lambda x. cut(x,cucumber) }\\
         & \mc{2}{\it \rightarrow divided(cucumber)}\\
\bapply{2}&\\
\mc{2}{\it AP}& \\
\mc{2}{\it cut(knife,cucumber)} & \\
\mc{2}{\it \rightarrow divided(cucumber)} &
}
\end{center}

\section{Learning Model and Semantic Parsing}
\label{sec:learning}

After having defined the formalism and application rule, instead of manually writing down all the possible CCG representations for each entity, we would like to apply a learning technique to derive them from the paired training corpus. Here we adopt the learning model of \cite{zettlemoyerlearning}, and use it to assign weights to the semantic representation of actions. Since an action may have multiple possible syntactic and semantic representations assigned to it, we use the probabilistic model to assign weights to these representations.

\subsection{Learning Approach}

First we assume that complete syntactic parses of the observed action are available, and in fact a manipulation action can have several different parses. The parsing uses a probabilistic combinatorial categorial grammar framework similar to the one given by \cite{zettlemoyer2007online}. We assume a probabilistic categorial grammar (PCCG) based on a log linear model. $M$ denotes a manipulation task, $L$ denotes the semantic representation of the task, and $T$ denotes its parse tree. The probability of a particular syntactic and semantic parse is given as:
\begin{eqnarray}
P(L,T|M; \Theta) = \frac{e^{f(L,T,M) \cdot \Theta}}{\sum_{(L,T)}e^{f(L,T,M) \cdot \Theta}}
\end{eqnarray}
where $f$ is a mapping of the triple ($L,T,M$) to feature vectors $\in R^d$, and the $\Theta \in R^d$ represents the weights to be learned. Here we use only lexical features, where each feature counts the number of times a lexical entry is used in $T$. Parsing a manipulation task under PCCG equates to finding $L$ such that $P(L|M;\Theta)$ is maximized:
\begin{eqnarray}
argmax_{L}P(L|M;\Theta) \nonumber \\ = argmax_{L}\sum_{T}P(L,T|M;\Theta).
\end{eqnarray}

We use dynamic programming techniques to calculate the most probable parse for the manipulation task. In this paper, the implementation from \cite{baral2011using} is adopted, where an inverse-$\lambda$ technique is used to generalize new semantic representations. The generalization of lexicon rules are essential for our system to deal with unknown actions presented during the testing phase.

\section{Experiments}

\subsection{Manipulation Action (MANIAC) Dataset}

\cite{aksoytamosiunaitewoergoetter2014} provides a manipulation action dataset with 8 different manipulation actions (cutting, chopping, stirring, putting, taking, hiding, uncovering, and pushing), each of which consists of 15 different versions performed by 5 different human actors\footnote{Dataset available for download at \url{https://fortknox.physik3.gwdg.de/cns/index.php?page=maniac-dataset}.}. There are in total 30 different objects manipulated in all demonstrations. All manipulations were recorded with the Microsoft Kinect sensor and serve as {\bf training} data here. 

The MANIAC data set contains another 20 long and complex chained manipulation sequences (e.g. ``making a sandwich'') which consist of a total of 103 different versions of these 8 manipulation tasks performed in different orders with novel objects under different circumstances. These serve as {\bf testing} data for our experiments. 

\cite{aksoytamosiunaitewoergoetter2014,aksoyCVIU2014} developed a semantic event chain based model free decomposition approach. It is an unsupervised probabilistic method that measures the frequency of the changes in the spatial relations embedded in event chains, in order to extract the subject and patient visual segments. It also decomposes the long chained complex testing actions into their primitive action components according to the spatio-temporal relations of the manipulator. Since the visual recognition is not the core of this work, we omit the details here and refer the interested reader to \cite{aksoytamosiunaitewoergoetter2014,aksoyCVIU2014}. All these features make the MANIAC dataset a great testing bed for both the theoretical framework and the implemented system presented in this work.  

\subsection{Training Corpus}

We first created a training corpus by annotating the 120 training clips from the MANIAC dataset, in the format of observed triplets (subject action patient) and a corresponding semantic representation of the action as well as its consequence. The semantic representations in $\lambda$-calculus format are given by human annotators after watching each action clip. A set of sample training pairs are given in Table.\ref{tb:tr_ex} (one from each action category in the training set). Since every training clip contains one single full execution of each manipulation action considered, the training corpus thus has a total of 120 paired training samples. 

\begin{table}[!ht]

\scriptsize
  \centering
  \begin{tabular}{ | m{.06\textwidth} | m{0.13\textwidth} | m{0.22\textwidth} | }
    \hline
    Snapshot & triplet & semantic representation \\ \hline
    \begin{minipage}{.07\textwidth}
      \includegraphics[width=\linewidth, height=10mm]{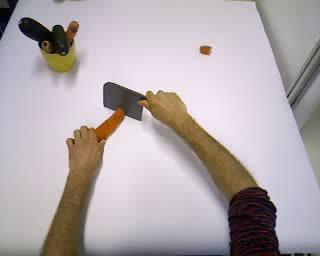}
    \end{minipage}
    &
    \pbox{0.15\textwidth}{cleaver chopping carrot}
    &
    \pbox{0.22\textwidth}{$chopping(cleaver,carrot)$ \\
    $\rightarrow divided(carrot)$}
     \\ \hline
     \begin{minipage}{.07\textwidth}
      \includegraphics[width=\linewidth, height=10mm]{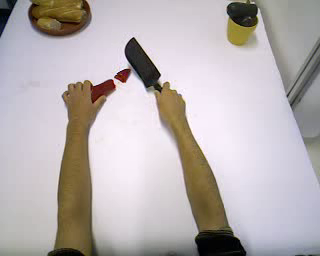}
    \end{minipage}
    &
    \pbox{0.15\textwidth}{spatula cutting pepper}
    &
    \pbox{0.22\textwidth}{$cutting(spatula,pepper)$ \\
    $\rightarrow divided(pepper)$}
     \\ \hline
     \begin{minipage}{.07\textwidth}
      \includegraphics[width=\linewidth, height=10mm]{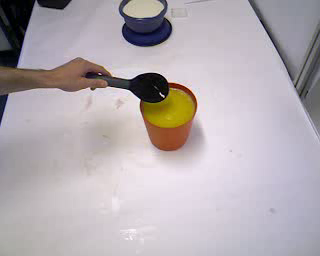}
    \end{minipage}
    &
    \pbox{0.15\textwidth}{spoon stirring bucket}
    &
    \pbox{0.22\textwidth}{$stirring(spoon,bucket)$}
     \\ \hline
     \begin{minipage}{.07\textwidth}
      \includegraphics[width=\linewidth, height=10mm]{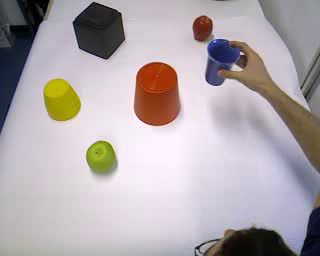}
    \end{minipage}
    &
    \pbox{0.15\textwidth}{cup take\_down bucket}
    &
    \pbox{0.22\textwidth}{$take\_down(cup,bucket)$ \\
    $\rightarrow \neg connected(cup,bucket)$ \\ 
    $ \wedge moved(cup)$}
     \\ \hline
     \begin{minipage}{.07\textwidth}
      \includegraphics[width=\linewidth, height=10mm]{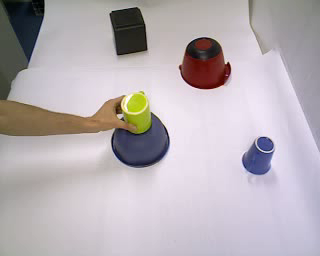}
    \end{minipage}
    &
    \pbox{0.15\textwidth}{cup put\_on\_top bowl}
    &
    \pbox{0.22\textwidth}{$put\_on\_top(cup,bowl)$ \\
    $\rightarrow on\_top(cup,bowl)$\\
    $\wedge moved(cup)$}
     \\ \hline
     \begin{minipage}{.07\textwidth}
      \includegraphics[width=\linewidth, height=10mm]{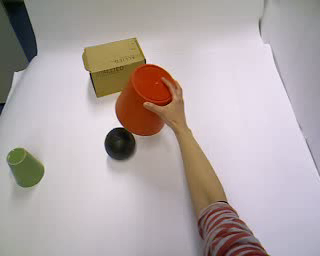}
    \end{minipage}
    &
    \pbox{0.15\textwidth}{bucket hiding ball}
    &
    \pbox{0.22\textwidth}{$hiding(bucket,ball)$ \\
    $\rightarrow contained(bucket,ball) $\\
    $\wedge moved(bucket)$}
     \\ \hline
     \begin{minipage}{.07\textwidth}
      \includegraphics[width=\linewidth, height=10mm]{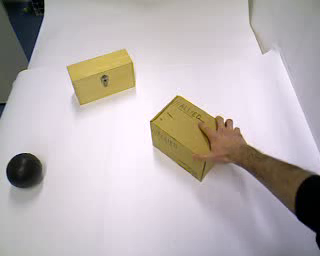}
    \end{minipage}
    &
    \pbox{0.15\textwidth}{hand pushing box}
    &
    \pbox{0.22\textwidth}{$pushing(hand,box)$ \\
    $\rightarrow moved(box)$}
     \\ \hline
     \begin{minipage}{.07\textwidth}
      \includegraphics[width=\linewidth, height=10mm]{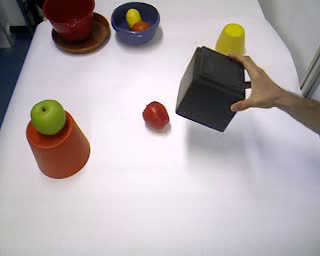}
    \end{minipage}
    &
    \pbox{0.15\textwidth}{box uncover apple}
    &
    \pbox{0.22\textwidth}{$uncover(box,apple)$ \\
    $\rightarrow appear(apple) $ \\
    $\wedge moved(box)$}
     \\ \hline
  \end{tabular}
  \caption{Example annotations from training corpus, one per manipulation action category.}\label{tb:tr_ex}
\end{table}

We also assume the system knows that every ``object'' involved in the corpus is an entity of its own type, for example:
\begin{smalign}
 Knife &:= N : knife \nonumber \\
 Bowl &:= N : bowl \nonumber \\
& ... ... \nonumber
\end{smalign}
Additionally, we assume the syntactic form of each ``action'' has a main type $(AP\backslash NP)/NP$ (see Sec.~\ref{sec:ccg}). These two sets of rules form the initial seed lexicon for learning.

\subsection{Learned Lexicon}
We applied the learning technique mentioned in Sec.~\ref{sec:learning}, and we used the NL2KR implementation from \cite{baral2011using}. The system learns and generalizes a set of lexicon entries (syntactic and semantic) for each action categories from the training corpus accompanied with a set of weights. We list the one with the largest weight for each action here respectively:
\begin{smalign}
Chopping := & (AP\backslash NP)/NP :  \lambda x. \lambda y. chopping(x,y)  \nonumber \\
&\rightarrow divided(y) \nonumber \\
Cutting := & (AP\backslash NP)/NP :  \lambda x. \lambda y. cutting(x,y) \nonumber \\
& \rightarrow divided(y) \nonumber \\
Stirring := & (AP\backslash NP)/NP :  \lambda x. \lambda y. stirring(x,y)  \nonumber \\
Take\_down := & (AP\backslash NP)/NP : \lambda x. \lambda y. take\_down(x,y)  \nonumber \\
& \rightarrow \neg connected(x,y) \wedge moved(x) \nonumber \\
Put\_on\_top := & (AP\backslash NP)/NP : \lambda x. \lambda y. put\_on\_top(x,y)  \nonumber \\
& \rightarrow on\_top(x,y) \wedge moved(x) \nonumber \\
Hiding := & (AP\backslash NP)/NP : \lambda x. \lambda y. hiding(x,y)  \nonumber \\
& \rightarrow contained(x,y) \wedge moved(x) \nonumber \\
Pushing := & (AP\backslash NP)/NP : \lambda x. \lambda y. pushing(x,y)  \nonumber \\
& \rightarrow moved(y) \nonumber \\
Uncover := & (AP\backslash NP)/NP : \lambda x. \lambda y. uncover(x,y)  \nonumber \\
& \rightarrow appear(y) \wedge moved(x) .\nonumber
\end{smalign}

The set of seed lexicon and the learned lexicon entries are further used to probabilistically parse the detected triplet sequences from the 20 long manipulation activities in the testing set. 

\subsection{Deducing Semantics}

Using the decomposition technique from \cite{aksoytamosiunaitewoergoetter2014,aksoyCVIU2014}, the reported system is able to detect a sequence of action triplets in the form of (Subject Action Patient) from each of the testing sequence in MANIAC dataset. 
Briefly speaking, the event chain
representation \cite{aksoy2011learning} of the observed long
manipulation activity is first scanned to estimate the main
manipulator, i.e. the hand, and manipulated objects, e.g. knife, in the scene 
without employing any visual feature-based object recognition method. 
Solely based on the interactions between the hand and
manipulated objects in the scene, the event chain
is partitioned into chunks. 
These chunks are further
fragmented into sub-units to detect parallel action streams. Each parsed Semantic Event Chain
(SEC) chunk is then compared with the model
SECs in the library to decide whether the current
SEC sample belongs to one of the known manipulation models or represents a novel manipulation.
SEC models, stored in
the library, are learned in an on-line unsupervised fashion
using the semantics of manipulations derived from a given
set of training data in order to create a large vocabulary of
single atomic manipulations.
%The framework is running in an automated and unsupervised manner to monitor chained manipulation sequences performed either sequentially or in
%parallel.

\begin{figure*}[!t]
\centering
\includegraphics[width=\textwidth]{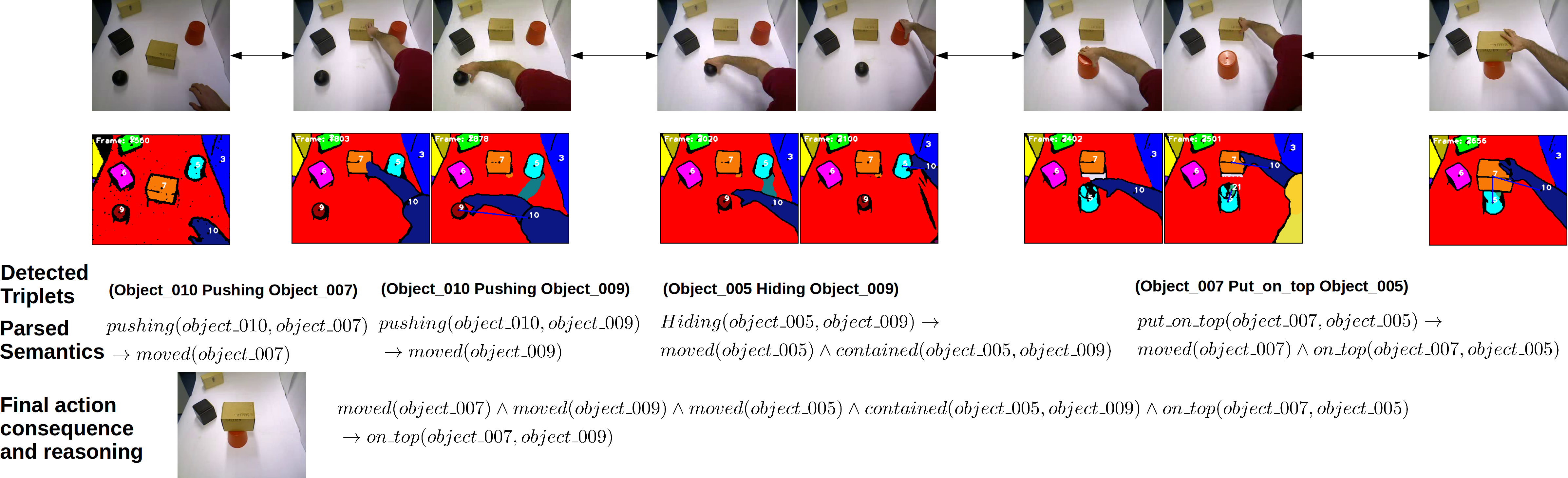}
\caption{System output on complex chained manipulation testing sequence one. The segmentation output and detected triplets are from \cite{aksoyCVIU2014}}.
\label{fig:test_ex1}
\end{figure*}

\begin{figure*}[!t]
\centering
\includegraphics[width=\textwidth]{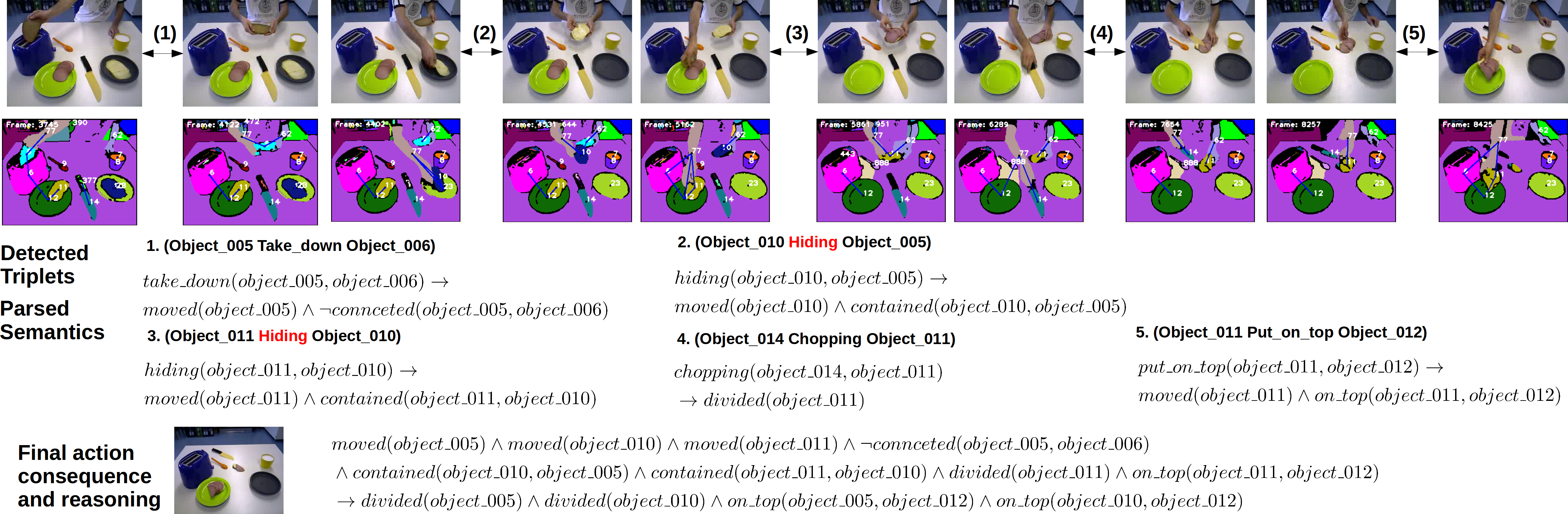}
\caption{System output on the 18th complex chained manipulation testing sequence. The segmentation output and detected triplets are from \cite{aksoyCVIU2014}}.
\label{fig:test_ex2}
\end{figure*}

For the different testing sequence, the number of triplets detected ranges from two to seven. In total, we are able to collect 90 testing detections and they serve as the testing corpus. However, since many of the objects used in the testing data are not present in the training set, an object model-free approach is adopted and thus ``subject'' and ``patient'' fields are filled with segment IDs instead of a specific object name. Fig.~\ref{fig:test_ex1} and ~\ref{fig:test_ex2} show several examples of the detected triplets accompanied with a set of key frames from the testing sequences. Nevertheless, the method we used here can 1) generalize the unknown segments into the category of object entities and 2) generalize the unknown actions (those that do not exist in the training corpus) into the category of action function. This is done by automatically generalizing the following two types of lexicon entries using the inverse-$\lambda$ technique from \cite{baral2011using}:
\begin{smalign}
 Object\_[ID] := & N : object\_[ID] \nonumber \\
 Unknown := & (AP\backslash NP)/NP : \lambda x. \lambda y. unknown(x,y)  \nonumber 
\end{smalign}

Among the $90$ detected triplets, using the learned lexicon we are able to parse all of them into semantic representations. Here we pick the representation with the highest probability after parsing as the individual action semantic representation. The ``parsed semantics'' rows of Fig.~\ref{fig:test_ex1} and ~\ref{fig:test_ex2} show several example action semantics on testing sequences. Taking the fourth sub-action from Fig.~\ref{fig:test_ex2} as an example, the visually detected triplets based on segmentation and spatial decomposition is $(Object\_014, Chopping, Object\_011)$. After semantic parsing, the system predicts that $divided(Object\_011)$. The complete training corpus and parsed results of the testing set will be made publicly available for future research.

\subsection{Reasoning Beyond Observations}

As mentioned before, because of  the use of  $\lambda$-calculus for representing action semantics, the obtained data can  naturally be used to do logical reasoning beyond observations. This by itself is a very interesting research topic and it is beyond this paper's scope. However by applying a couple of common sense Axioms on the testing data, we can provide some flavor of this idea.

\textbf{Case study one: }See the ``final action consequence and reasoning'' row of Fig.~\ref{fig:test_ex1} for case one. Using propositional logic and axiom schema, we can represent the common sense statement (``if an object $x$ is contained in object $y$, and object $z$ is on top of object $y$, then object $z$ is on top of object $x$'') as follows:

\textbf{Axiom (1): }$\exists x,y,z, contained(y,x) \wedge on\_top(z,y) \rightarrow on\_top(z,x)$.  

Then it is trivial to deduce an additional final action consequence in this scenario that ($on\_top(object\_007,object\_009)$). This matches the fact: the yellow box which is put on top of the red bucket is also on top of the black ball. 

\textbf{Case study two: }  See the ``final action consequence and reasoning'' row of Fig.~\ref{fig:test_ex2} for a more complicated case. Using propositional logic and axiom schema, we can represent three common sense statements: 

1) ``if an object $y$ is contained in object $x$, and object $z$ is contained in object $y$, then object $z$ is contained in object $x$'';

2) ``if an object $x$ is contained in object $y$, and object $y$ is divided, then object $x$ is divided''; 

3) ``if an object $x$ is contained in object $y$, and object $y$ is on top of object $z$, then object $x$ is on top of object $z$'' as follows:

\textbf{Axiom (2): }$\exists x,y,z, contained(y,x) \wedge contained(z,y) \rightarrow contained(z,x)$.  

\textbf{Axiom (3): }$\exists x,y, contained(y,x) \wedge divided(y) \rightarrow divided(x)$.  

\textbf{Axiom (4): }$\exists x,y,z, contained(y,x) \wedge on\_top(y,z) \rightarrow on\_top(x,z)$.

With these common sense Axioms, the system is able to deduce several additional final action consequences in this scenario: 
\begin{smalign}
&divided(object\_005) \wedge  divided(object\_010) \nonumber \\
&\wedge on\_top(object\_005, object\_012) \nonumber \\
&\wedge on\_top(object\_010, object\_012). \nonumber
\end{smalign}

From Fig.~\ref{fig:test_ex2}, we can see that these additional consequences indeed match the facts: 1) the bread and cheese which are covered by ham are also divided, even though from observation the system only detected the ham being cut; 2) the divided bread and cheese are also on top of the plate, even though from observation the system only detected the ham being put on top of the plate.   

We applied the four Axioms on the 20 testing  action sequences and deduced the ``hidden'' consequences from observation. To evaluate our system performance quantitatively, we first annotated all the final action consequences (both obvious and ``hidden'' ones) from the 20 testing sequences as ground-truth facts. In total there are 122 consequences annotated. Using perception only \cite{aksoyCVIU2014}, due to the decomposition errors (such as the red font ones in Fig.~\ref{fig:test_ex2}) the system can detect 91 consequences correctly, yielding a 74\% correct rate. After applying the four Axioms and reasoning, our system is able to detect 105 consequences correctly, yielding a 86\% correct rate. Overall, this  is a 15.4\% of improvement.

Here we want to mention a caveat: there are definitely other common sense Axioms that we are not able to address  in the current implementation. However, from the case studies presented, we can see that using the presented formal framework, our system is able to reason about manipulation action goals instead of just observing what is happening visually. This capability is essential for intelligent agents to imitate action goals from observation.

\section{Conclusion and Future Work}

In this paper we presented a formal computational framework for modeling manipulation actions based on a Combinatory Categorial Grammar. An empirical study on a large manipulation action dataset validates that 1) with the introduced formalism, a learning system can be devised to deduce the semantic meaning of manipulation actions in $\lambda$-schema; 2) with the learned schema and several common sense Axioms, our system is able to reason beyond just observation and deduce ``hidden'' action consequences, yielding a decent performance improvement. 

Due to the limitation of current testing scenarios, we conducted experiments only considering a relatively small set of seed lexicon rules and logical expressions. Nevertheless, we want to mention that the presented CCG framework can also be extended to learn the formal logic representation of more complex manipulation action semantics. For example, the temporal order of manipulation actions can be modeled by considering a seed rule such as $AP\backslash AP : \lambda f. \lambda g. before(f(\cdot),g(\cdot))$, where $before(\cdot,\cdot)$ is a temporal predicate. For actions in this paper we consider seed main type $(AP\bs NP)/NP$. For more general manipulation scenarios, based on whether the action is transitive or intransitive, the main types of action can be extended to include  $AP\bs NP$. 

Moreover, the logical expressions can also be extended to include universal quantification $\forall$ and existential quantification $\exists$. Thus, manipulation action such as ``knife cut every tomato'' can be parsed into a representation as $\forall x. tomato(x) \wedge cut(knife,x) \rightarrow divided(x)$ (the parse is given in the following chart). Here, the concept ``every'' has a main type of $NP\backslash NP$ and semantic meaning of $\forall x. f(x)$. The same framework can also extended to have other combinatory rules such as {\bf composition} and {\bf type-raising} \cite{steedman2002plans}. These are parts of the future work along the line of the presented work. 

\begin{center}
\small
\deriv{4}{
{\rm Knife}&{\rm Cut}& {\rm every} & {\rm Tomato}\\
\uline{1}& \uline{1}  &\uline{1}  & \uline{1}\\
\it N   & & &\it  N\\
\uline{1}& \uline{1} &\uline{1}   & \uline{1}\\
\it NP   &\it (AP\bs NP)/NP & NP\bs NP &\it  NP\\
\it knife   &\it \lambda x. \lambda y. cut(x,y) &\it \forall x. f(x) &\it  tomato\\
\it knife   &\it \rightarrow divided(y) &\it \forall x. f(x) &\it  tomato\\
         & &     \fapply{2}\\
         & & \mc{2}{NP} \\
         & & \mc{2}{\it \forall x. tomato(x)} \\
         &      \fapply{3}\\
         &  \mc{3}{\it AP\bs NP}\\
         &  \mc{3}{\it \forall y. \lambda x. tomato(y) \wedge cut(x,y) \rightarrow divided(y)}\\
\bapply{4} \\
\mc{4}{\it AP}  \\
\mc{4}{\it \forall y. tomato(y) \wedge cut(knife,y) \rightarrow divided(y)}
}
\end{center}

The presented computational linguistic framework enables an intelligent agent to predict and reason action goals from observation, and thus has many potential applications such as human intention prediction, robot action policy planning, human robot collaboration etc. We believe that our formalism of manipulation actions bridges computational linguistics, vision and robotics, and opens further research in Artificial Intelligence and Robotics. As the robotics industry is moving towards robots that function safely, effectively and autonomously to perform tasks in real-world unstructured environments, they will need to be able to understand the meaning of actions and acquire human-like common-sense reasoning capabilities.

\section{Acknowledgements} 

This research was funded in part by the support of the European Union under the Cognitive Systems program (project POETICON++), the National Science Foundation under  INSPIRE grant SMA 1248056, and by DARPA through U.S. Army grant W911NF-14-1-0384 under the Project:  Shared Perception, Cognition and Reasoning for Autonomy.

\newpage 

\bibliographystyle{acl}
\bibliography{format}

\begin{thebibliography}{}

\bibitem[\protect\citename{Aksoy and W{\"o}rg{\"o}tter}2015]{aksoyCVIU2014}
E~E. Aksoy and F.~W{\"o}rg{\"o}tter.
\newblock 2015.
\newblock Semantic decomposition and recognition of long and complex
  manipulation action sequences.
\newblock {\em International Journal of Computer Vision}, page Under Review.

\bibitem[\protect\citename{Aksoy \bgroup et al.\egroup
  }2011]{aksoy2011learning}
E.E. Aksoy, A.~Abramov, J.~D{\"o}rr, K.~Ning, B.~Dellen, and
  F.~W{\"o}rg{\"o}tter.
\newblock 2011.
\newblock Learning the semantics of object--action relations by observation.
\newblock {\em The International Journal of Robotics Research},
  30(10):1229--1249.

\bibitem[\protect\citename{Aksoy \bgroup et al.\egroup
  }2014]{aksoytamosiunaitewoergoetter2014}
E~E. Aksoy, M.~Tamosiunaite, and F.~W{\"o}rg{\"o}tter.
\newblock 2014.
\newblock Model-free incremental learning of the semantics of manipulation
  actions.
\newblock {\em Robotics and Autonomous Systems}, pages 1--42.

\bibitem[\protect\citename{Baral \bgroup et al.\egroup }2011]{baral2011using}
Chitta Baral, Juraj Dzifcak, Marcos~Alvarez Gonzalez, and Jiayu Zhou.
\newblock 2011.
\newblock Using inverse $\lambda$ and generalization to translate english to
  formal languages.
\newblock In {\em Proceedings of the Ninth International Conference on
  Computational Semantics}, pages 35--44. Association for Computational
  Linguistics.

\bibitem[\protect\citename{Ben-Arie \bgroup et al.\egroup }2002]{ben2002human}
Jezekiel Ben-Arie, Zhiqian Wang, Purvin Pandit, and Shyamsundar Rajaram.
\newblock 2002.
\newblock Human activity recognition using multidimensional indexing.
\newblock {\em Pattern Analysis and Machine Intelligence, IEEE Transactions
  on}, 24(8):1091--1104.

\bibitem[\protect\citename{Brand}1996]{brand1996understanding}
Matthew Brand.
\newblock 1996.
\newblock Understanding manipulation in video.
\newblock In {\em Proceedings of the Second International Conference on
  Automatic Face and Gesture Recognition}, pages 94--99, Killington,VT. IEEE.

\bibitem[\protect\citename{Chaudhry \bgroup et al.\egroup
  }2009]{chaudhry2009histograms}
R.~Chaudhry, A.~Ravichandran, G.~Hager, and R.~Vidal.
\newblock 2009.
\newblock Histograms of oriented optical flow and binet-cauchy kernels on
  nonlinear dynamical systems for the recognition of human actions.
\newblock In {\em Proceedings of the 2009 IEEE Intenational Conference on
  Computer Vision and Pattern Recognition}, pages 1932--1939, Miami,FL. IEEE.

\bibitem[\protect\citename{Chomsky}1957]{Chomsky1957}
N.~Chomsky.
\newblock 1957.
\newblock {\em Syntactic Structures}.
\newblock Mouton de Gruyter.

\bibitem[\protect\citename{Chomsky}1993]{chomsky1993lectures}
Noam Chomsky.
\newblock 1993.
\newblock {\em Lectures on government and binding: The {P}isa lectures}.
\newblock Walter de Gruyter.

\bibitem[\protect\citename{Gazzola \bgroup et al.\egroup
  }2007]{gazzola2007anthropomorphic}
V~Gazzola, G~Rizzolatti, B~Wicker, and C~Keysers.
\newblock 2007.
\newblock The anthropomorphic brain: the mirror neuron system responds to human
  and robotic actions.
\newblock {\em Neuroimage}, 35(4):1674--1684.

\bibitem[\protect\citename{Guha \bgroup et al.\egroup }2013]{guha2013plan}
Anupam Guha, Yezhou Yang, Cornelia Ferm{\"u}ller, and Yiannis Aloimonos.
\newblock 2013.
\newblock Minimalist plans for interpreting manipulation actions.
\newblock {\em Proceedings of the 2013 IEEE/RSJ International Conference on
  Intelligent Robots and Systems}, pages 5908--5914.

\bibitem[\protect\citename{Ivanov and Bobick}2000]{ivanov2000recognition}
Yuri~A. Ivanov and Aaron~F. Bobick.
\newblock 2000.
\newblock Recognition of visual activities and interactions by stochastic
  parsing.
\newblock {\em IEEE Transactions on Pattern Analysis and Machine Intelligence},
  22(8):852--872.

\bibitem[\protect\citename{Joo \bgroup et al.\egroup }2014]{joo2014visual}
Jungseock Joo, Weixin Li, Francis~F Steen, and Song-Chun Zhu.
\newblock 2014.
\newblock Visual persuasion: Inferring communicative intents of images.
\newblock In {\em Computer Vision and Pattern Recognition (CVPR), 2014 IEEE
  Conference on}, pages 216--223. IEEE.

\bibitem[\protect\citename{Kale \bgroup et al.\egroup
  }2004]{kale2004identification}
A.~Kale, A.~Sundaresan, AN~Rajagopalan, N.P. Cuntoor, A.K. Roy-Chowdhury,
  V.~Kruger, and R.~Chellappa.
\newblock 2004.
\newblock Identification of humans using gait.
\newblock {\em IEEE Transactions on Image Processing}, 13(9):1163--1173.

\bibitem[\protect\citename{Kuehne \bgroup et al.\egroup
  }2014]{kuehne2014language}
Hilde Kuehne, Ali Arslan, and Thomas Serre.
\newblock 2014.
\newblock The language of actions: Recovering the syntax and semantics of
  goal-directed human activities.
\newblock In {\em Computer Vision and Pattern Recognition (CVPR), 2014 IEEE
  Conference on}, pages 780--787. IEEE.

\bibitem[\protect\citename{Matuszek \bgroup et al.\egroup
  }2011]{matuszek2012joint}
Cynthia Matuszek, Nicholas FitzGerald, Luke Zettlemoyer, Liefeng Bo, and Dieter
  Fox.
\newblock 2011.
\newblock A joint model of language and perception for grounded attribute
  learning.
\newblock In {\em International Conference on Machine learning (ICML)}.

\bibitem[\protect\citename{Matuszek \bgroup et al.\egroup
  }2014]{matuszek2014learning}
Cynthia Matuszek, Liefeng Bo, Luke Zettlemoyer, and Dieter Fox.
\newblock 2014.
\newblock Learning from unscripted deictic gesture and language for human-robot
  interactions.
\newblock In {\em Twenty-Eighth AAAI Conference on Artificial Intelligence}.

\bibitem[\protect\citename{Moeslund \bgroup et al.\egroup
  }2006]{moeslund2006survey}
T.B. Moeslund, A.~Hilton, and V.~Kr{\"u}ger.
\newblock 2006.
\newblock A survey of advances in vision-based human motion capture and
  analysis.
\newblock {\em Computer vision and image understanding}, 104(2):90--126.

\bibitem[\protect\citename{Mooney}2008]{mooney2008learning}
Raymond~J Mooney.
\newblock 2008.
\newblock Learning to connect language and perception.
\newblock In {\em AAAI}, pages 1598--1601.

\bibitem[\protect\citename{Moore and Essa}2002]{moore2002recognizing}
Darnell Moore and Irfan Essa.
\newblock 2002.
\newblock Recognizing multitasked activities from video using stochastic
  context-free grammar.
\newblock In {\em Proceedings of the National Conference on Artificial
  Intelligence}, pages 770--776, Menlo Park, CA. AAAI.

\bibitem[\protect\citename{Pastra and Aloimonos}2012]{pastra2012minimalist}
K.~Pastra and Y.~Aloimonos.
\newblock 2012.
\newblock The minimalist grammar of action.
\newblock {\em Philosophical Transactions of the Royal Society B: Biological
  Sciences}, 367(1585):103--117.

\bibitem[\protect\citename{Pirsiavash \bgroup et al.\egroup
  }2014]{pirsiavash2014inferring}
Hamed Pirsiavash, Carl Vondrick, and Antonio Torralba.
\newblock 2014.
\newblock Inferring the why in images.
\newblock {\em arXiv preprint arXiv:1406.5472}.

\bibitem[\protect\citename{Rizzolatti \bgroup et al.\egroup
  }2001]{rizzolatti2001neurophysiological}
Giacomo Rizzolatti, Leonardo Fogassi, and Vittorio Gallese.
\newblock 2001.
\newblock Neurophysiological mechanisms underlying the understanding and
  imitation of action.
\newblock {\em Nature Reviews Neuroscience}, 2(9):661--670.

\bibitem[\protect\citename{Ryoo and Aggarwal}2006]{ryoo2006recognition}
Michael~S Ryoo and Jake~K Aggarwal.
\newblock 2006.
\newblock Recognition of composite human activities through context-free
  grammar based representation.
\newblock In {\em Proceedings of the 2006 IEEE Conference on Computer Vision
  and Pattern Recognition}, volume~2, pages 1709--1718, New York City, NY.
  IEEE.

\bibitem[\protect\citename{Saisan \bgroup et al.\egroup
  }2001]{saisan2001dynamic}
P.~Saisan, G.~Doretto, Y.N. Wu, and S.~Soatto.
\newblock 2001.
\newblock Dynamic texture recognition.
\newblock In {\em Proceedings of the 2001 IEEE Intenational Conference on
  Computer Vision and Pattern Recognition}, volume~2, pages 58--63, Kauai, HI.
  IEEE.

\bibitem[\protect\citename{Steedman}2000]{steedman2000syntactic}
Mark Steedman.
\newblock 2000.
\newblock {\em The syntactic process}, volume~35.
\newblock MIT Press.

\bibitem[\protect\citename{Steedman}2002]{steedman2002plans}
Mark Steedman.
\newblock 2002.
\newblock Plans, affordances, and combinatory grammar.
\newblock {\em Linguistics and Philosophy}, 25(5-6):723--753.

\bibitem[\protect\citename{Summers-Stay \bgroup et al.\egroup
  }2013]{summersusing}
D.~Summers-Stay, C.L. Teo, Y.~Yang, C.~Ferm{\"u}ller, and Y.~Aloimonos.
\newblock 2013.
\newblock Using a minimal action grammar for activity understanding in the real
  world.
\newblock In {\em Proceedings of the 2013 IEEE/RSJ International Conference on
  Intelligent Robots and Systems}, pages 4104--4111, Vilamoura, Portugal. IEEE.

\bibitem[\protect\citename{Tellex \bgroup et al.\egroup
  }2014]{tellex2014learning}
Stefanie Tellex, Pratiksha Thaker, Joshua Joseph, and Nicholas Roy.
\newblock 2014.
\newblock Learning perceptually grounded word meanings from unaligned parallel
  data.
\newblock {\em Machine Learning}, 94(2):151--167.

\bibitem[\protect\citename{Turaga \bgroup et al.\egroup
  }2008]{turaga2008machine}
P.~Turaga, R.~Chellappa, V.S. Subrahmanian, and O.~Udrea.
\newblock 2008.
\newblock Machine recognition of human activities: A survey.
\newblock {\em IEEE Transactions on Circuits and Systems for Video Technology},
  18(11):1473--1488.

\bibitem[\protect\citename{Xie \bgroup et al.\egroup }2013]{xie2013inferring}
Dan Xie, Sinisa Todorovic, and Song-Chun Zhu.
\newblock 2013.
\newblock Inferring ``dark matter'' and ``dark energy'' from videos.
\newblock In {\em Computer Vision (ICCV), 2013 IEEE International Conference
  on}, pages 2224--2231. IEEE.

\bibitem[\protect\citename{Yang \bgroup et al.\egroup }2013]{yezhou2013cvpr}
Yezhou Yang, Cornelia Ferm{\"u}ller, and Yiannis Aloimonos.
\newblock 2013.
\newblock Detection of manipulation action consequences ({M}{A}{C}).
\newblock In {\em Proceedings of the 2013 IEEE Conference on Computer Vision
  and Pattern Recognition}, pages 2563--2570, Portland, OR. IEEE.

\bibitem[\protect\citename{Yang \bgroup et al.\egroup }2014]{YezhouACS}
Y.~Yang, A.~Guha, C.~Fermuller, and Y.~Aloimonos.
\newblock 2014.
\newblock A cognitive system for understanding human manipulation actions.
\newblock {\em Advances in Cognitive Sysytems}, 3:67--86.

\bibitem[\protect\citename{Yang \bgroup et al.\egroup }2015]{yang2015robot}
Yezhou Yang, Yi~Li, Cornelia Fermuller, and Yiannis Aloimonos.
\newblock 2015.
\newblock Robot learning manipulation action plans by ``watching''
  unconstrained videos from the world wide web.
\newblock In {\em The Twenty-Ninth AAAI Conference on Artificial Intelligence
  (AAAI-15)}.

\bibitem[\protect\citename{Yilmaz and Shah}2005]{yilmaz2005actions}
A.~Yilmaz and M.~Shah.
\newblock 2005.
\newblock Actions sketch: A novel action representation.
\newblock In {\em Proceedings of the 2005 IEEE Intenational Conference on
  Computer Vision and Pattern Recognition}, volume~1, pages 984--989, San
  Diego, CA. IEEE.

\bibitem[\protect\citename{Zettlemoyer and Collins}2005]{zettlemoyerlearning}
Luke~S Zettlemoyer and Michael Collins.
\newblock 2005.
\newblock Learning to map sentences to logical form: Structured classification
  with probabilistic categorial grammars.
\newblock In {\em UAI}.

\bibitem[\protect\citename{Zettlemoyer and Collins}2007]{zettlemoyer2007online}
Luke~S Zettlemoyer and Michael Collins.
\newblock 2007.
\newblock Online learning of relaxed ccg grammars for parsing to logical form.
\newblock In {\em EMNLP-CoNLL}, pages 678--687.

\end{thebibliography}

\end{document}